\pdfoutput=1

\documentclass[11pt]{article}

\usepackage[preprint]{latex/acl}

\PassOptionsToPackage{dvipsnames}{xcolor}
\usepackage{times}
\usepackage{latexsym}
\usepackage{amsmath}
\usepackage{amssymb}
\usepackage{booktabs}
\usepackage{todonotes}
\usepackage{enumitem}
\usepackage{multicol}
\usepackage{subcaption}
\usepackage{xcolor}
\usepackage[utf8]{inputenc}
\usepackage{cjhebrew}

\usepackage{soul}
\usepackage{tcolorbox}
\definecolor{activation_highlight}{RGB}{208,96,40}
\definecolor{ActivationBackground}{RGB}{252,252,251}
\definecolor{ActivationBorder}{RGB}{246,246,243}
\definecolor{ActivationBorderIntense}{RGB}{176,176,143}

\definecolor{contrast_light_blue}{HTML}{4BABE1}
\definecolor{contrast_orange}{HTML}{EE8C44}
\definecolor{intervention_highlight}{HTML}{EED8D8}
\definecolor{intervention_color_1}{HTML}{B270AF}

\definecolor{intervention_color_2}{HTML}{6FBFAA}
\definecolor{intervention_color_3}{HTML}{4BABE0}

\usepackage{lipsum}

\newcommand\blfootnote[1]{%
  \begingroup
  \renewcommand\thefootnote{}\footnote{#1}%
  \addtocounter{footnote}{-1}%
  \endgroup
}

\usepackage{adjustbox}

\definecolor{redhighlight}{HTML}{d7a3a3}
\definecolor{ColorEN}{RGB}{169,55,19}
\definecolor{ColorFR}{RGB}{253,140,110}
\definecolor{ColorDE}{RGB}{123,85,79}
\definecolor{ColorNL}{RGB}{242,205,185}
\definecolor{ColorFI}{RGB}{236,16,47}
\definecolor{ColorTR}{RGB}{144,45,84}
\definecolor{ColorZH}{RGB}{248,35,135}
\definecolor{ColorJA}{RGB}{246,189,83}
\definecolor{ColorID}{RGB}{156,111,8}

\usepackage{tikz}
\usepackage{pgfplots}
\pgfplotsset{compat=1.18}
\usetikzlibrary{positioning}

\usepackage[T1]{fontenc}

\usepackage[utf8]{inputenc}

\usepackage{microtype}

\usepackage{inconsolata}

\usepackage{graphicx}

\definecolor{blue}{HTML}{3E74D1}
\definecolor{red}{HTML}{E22146}
\definecolor{green}{HTML}{70b642}
\definecolor{violet}{HTML}{af649b}
\definecolor{lightgray}{HTML}{c6c6c6}

\usepackage{amsmath,amsfonts,bm}

\def\eqref#1{equation~\ref{#1}}

\def\1{\bm{1}}

\def\va{{\mathbf{a}}}
\def\vb{{\mathbf{b}}}

\def\vv{{\mathbf{v}}}

\def\vx{{\mathbf{x}}}

\def\vepsilon{{\bm\epsilon}}

\DeclareMathAlphabet{\mathsfit}{\encodingdefault}{\sfdefault}{m}{sl}
\SetMathAlphabet{\mathsfit}{bold}{\encodingdefault}{\sfdefault}{bx}{n}

\def\sR{{\mathbb{R}}}

\makeatletter
\def\@makefnmark{\smash{\hbox{\@textsuperscript{\normalfont\@thefnmark}}}}
\makeatother

\usepackage{pgfplots}
\usepackage{tikz}

\title{Large Language Models Share Representations of Latent Grammatical Concepts Across Typologically Diverse Languages}

\author{Jannik Brinkmann$^{1,2}$ \; \; Chris Wendler$^{3}$ \; \; Christian Bartelt$^{2}$\; \; Aaron Mueller$^{1,4}$ \\
$^1$Northeastern University\; $^2$TU Clausthal\; $^3$EPFL\; $^4$Technion -- IIT}

\begin{document}
\maketitle
\begin{abstract}
Human bilinguals often use similar brain regions to process multiple languages, depending on when they learned their second language and their proficiency. In large language models (LLMs), how are multiple languages learned and encoded?
In this work, we explore the extent to which LLMs share representations of morphsyntactic concepts such as grammatical number, gender, and tense across languages. 
We train sparse autoencoders on Llama-3-8B and Aya-23-8B, and demonstrate that abstract grammatical concepts are often encoded in feature directions shared across many languages. 
We use causal interventions to verify the multilingual nature of these representations; specifically, we show that ablating only multilingual features decreases classifier performance to near-chance across languages.
We then use these features to precisely modify model behavior in a machine translation task; this demonstrates both the generality and selectivity of these feature's roles in the network.
Our findings suggest that even models trained predominantly on English data can develop robust, cross-lingual abstractions of morphosyntactic concepts.
\end{abstract}

\section{Introduction}
In the brains of human bilinguals, syntax processing may occur in similar regions for their first and second language, depending on factors like when the second language was learned \citep{cargnelutti2019language}, language proficiency \citep{POLCZYNSKA20211}, among many other factors \citep{sulpizio2020bilingual,Costa2014}. In multilingual language models~\citep[LMs;][]{shannon}, 
how apt is the analogy of shared processing to human bilinguals? If we desire parameter-efficiency, we might want multilingual representations of concepts such as grammatical number, rather than many language-specific representations of the same concept.
\blfootnote{We release code, data, and autoencoders at \nolinkurl{github.com/jannik-brinkmann/multilingual-features}.}

\begin{figure}[t!]
    \centering
    \includegraphics[width=1\columnwidth]{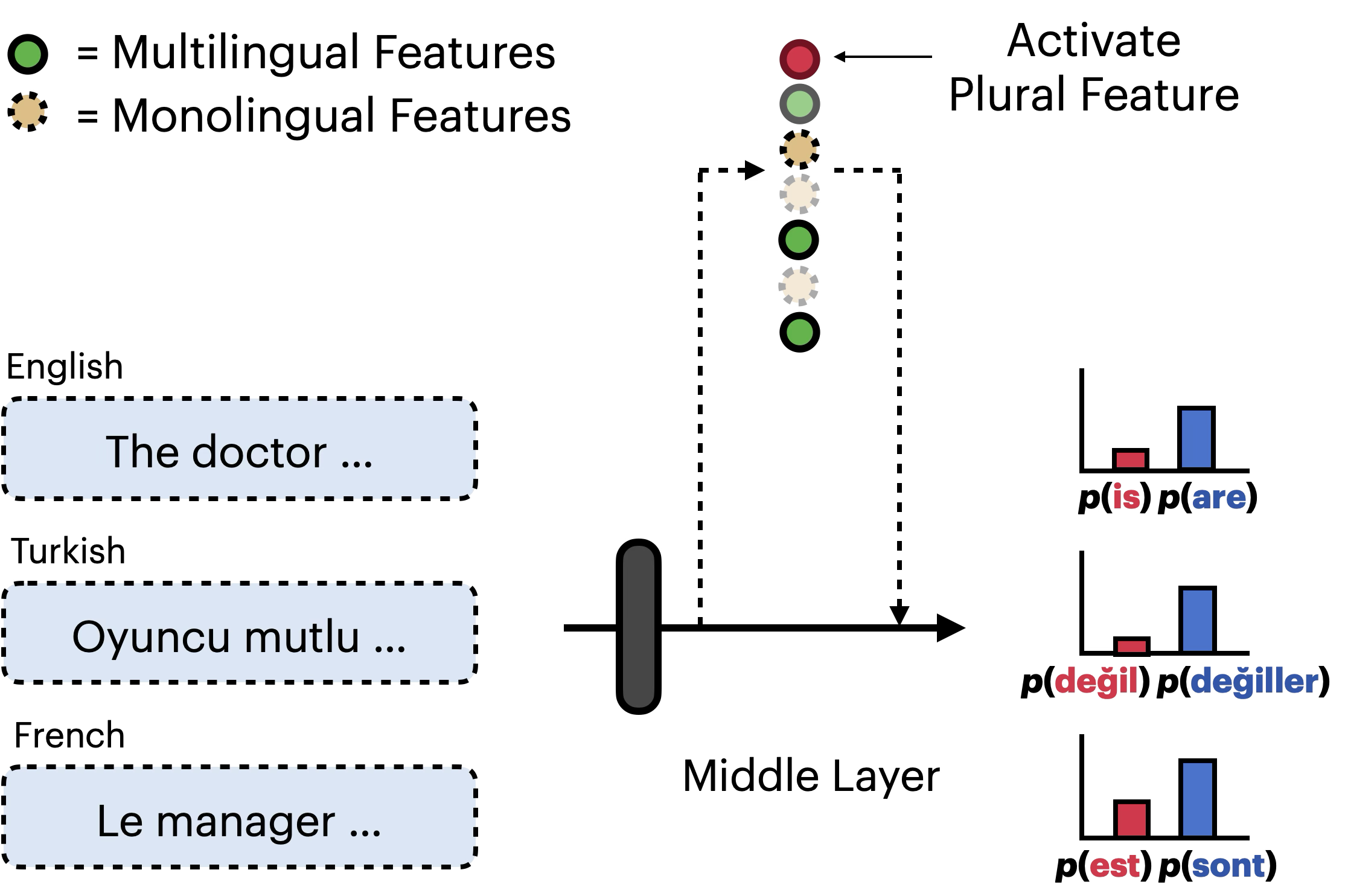}
    \caption{Using sparse autoencoders, we find that language models share representations of grammatical concepts across languages. 
    By intervening on these multilingual representations, we can change the model behavior given inputs in different languages. 
    For example, we can make the model predict plural verbs in \textit{different} languages by activating the \textit{same} plural feature.}
    \label{fig:enter-label}
\end{figure}

Past work has emphasized language-balanced pretraining corpora~\citep[e.g.,][]{conneau-etal-2020-unsupervised, xue-etal-2021-mt5, mohammadshahi-etal-2022-small}, such that an LM could be said to have many primary languages. However, many of the best-performing multilingual LMs are now primarily English models, trained on over 90\% English text~\citep[e.g.,][]{meta_llama_3}. Why do these models perform so well in non-English languages? We hypothesize that these models learn \textbf{generalizable abstractions} that enable more efficient learning of new languages. In other words, being able to deeply characterize a smaller distribution could allow models to acquire more robust abstractions that may generalize more effectively to a wider distribution post-hoc; in contrast, more balanced corpora (wider distributions) could encourage the model to start by learning language-specific abstractions, which are potentially never merged into higher-level language-invariant concepts. For small models, on a synthetic language pair (English and a token-level duplication of it), \citet{schafer2024language} show that imbalanced datasets lead to higher sample efficiency and higher overall performance in both languages in the low-data regime.\footnote{For real languages, we expect that features will be shared across languages to a much greater extent.}

In large language models, multilinguality has both practical and theoretical scientific value. High performance on natural language processing tasks across many languages increases the impact and inclusivity of language technologies, and in a more compute-efficient manner than would be possible by training a large series of monolingual models. Moreover, multilingual models are hypothesized to be able to outperform monolingual models in low-resource languages: past work has emphasized the importance of \textbf{cross-lingual transfer}, where knowledge in one language is shared with another language whose corpora did not contain the information of interest~\citep{libovicky-etal-2020-language, chang-etal-2022-geometry, hua-etal-2024-mothello}. More abstractly, the extent to which abstract concepts are shared across languages addresses a key question: what kinds of multi-lingual processing mechanisms are acquirable from exposure to large-scale text distributions? It would be more parameter-efficient to learn generalizable abstractions that apply across languages, as opposed to redundantly learning the same grammatical feature for each language separately. Moreover, the existence of generalizable abstractions informs debates on memorization versus generalization: cross-linguistic generalization suggests a more sophisticated and broad application of particular concepts.

Past work has largely investigated multilingual language models using behavioral/benchmarking analyses~\citep[\textit{inter alia}]{gupta-srikumar-2021-x, raganato-etal-2020-xl} or neuron-level mechanistic analyses~\citep{stanczak-etal-2022-neurons, varda-marelli-2023-data}. \citeauthor{stanczak-etal-2022-neurons} observe that masked language models often share neurons across languages for a particular concept, but causally verifying this is difficult, as counterfactual interventions to neurons often affect the processing of irrelevant concepts (see \S\ref{cha:methods}). To overcome this challenge, we make our units of causal analysis \emph{sparse autoencoder features}, which have been shown to be more \emph{monosemantic}, and therefore more human-interpretable~\citep{huben2024sparse, bricken2023monosemanticity, rajamanoharan2024improvingdictionarylearninggated}. This enables more precise causal interventions \citep{marks2024sparsefeaturecircuitsdiscovering,mueller2024questrightmediatorhistory}, and allows us to qualitatively verify that features of interest are truly sensitive to the hypothesized feature, rather than a spurious but close-enough feature.

In this work, we train a set of sparse autoencoders on the intermediate activations of Llama-3-8B~\cite{meta_llama_3} and Aya-3-8B~\citep{aryabumi2024aya} and locate (massively) multilingual features for various morphosyntactic concepts. 
We design experiments to quantify the degree to which these concepts are shared across languages, and validate their role in the models generations. 
Our results reveal that language models share morphosyntactic concept representations across typologically diverse languages, and that the internal \emph{lingua franca} of large language models may not be English words \emph{per se}, but rather concepts.\footnote{Though we note that these concepts are likely to be biased toward English-like representations, as discussed in \citet{wendler2024llamasworkenglishlatent}.}   
\section{Background}
\label{cha:methods}

\paragraph{Feature disentanglement using sparse autoencoders.}
The features underlying model computation are not guaranteed to be aligned to neuron bases; they may instead be represented in a distributed manner, such that there is a many-to-many relationship between neurons and concepts~\citep{hinton1986distributed,smolensky1986distributed}. In practice, single neurons\footnote{We use ``neuron'' to refer to a single dimension of any hidden representation vector in a model.} are often \emph{polysemantic}; in other words, they activate on a range of seemingly unrelated concepts~\citep{bolukbasi2021interpretabilityillusionbert, elhage2022superposition}.
For example,~\citet{bricken2023monosemanticity} observe that a single neuron in a small LM responds to a mixture of academic citations, English dialogue, HTTP requests, and Korean text. 
To address this, \citet{huben2024sparse, bricken2023monosemanticity} propose sparse autoencoders (SAEs) as a scalable technique for unsupervised discovery of interpretable feature directions in neural networks. 
Given an activation vector \smash{$\vx\in \sR^{d_\text{model}}$}, the autoencoder computes a decomposition 
\begin{equation}\label{eq:SAE-decomp}
    \vx = \hat{\vx} + \vepsilon(x) = \vb + \sum_{i}^{} f_i(\vx)\vv_i + \vepsilon(x)
\end{equation}
into an approximate reconstruction $\hat{\vx}$ as a linear combination of features $\vv_i$.
The features~$\vv_i\in \sR^{d_\text{model}}$ are unit vectors, the feature activations $f_i(\vx)\in \sR$ are a sparse set of coefficients, $\vb\in \sR^{d_\text{model}}$ is a bias term, and $\vepsilon(x) \in \sR^{d_{\text{model}}}$ is the approximation error. 

\citet{bricken2023monosemanticity} train SAEs by minimizing an L2 reconstruction error and an L1 regularization term to promote sparsity. 
However, the L1 regularization term introduces biases that can harm the accuracy of the reconstruction.
Therefore, we use the Gated SAE architecture~\citep{rajamanoharan2024improvingdictionarylearninggated}, which separates the functionality of selecting the features to use and estimating the activation magnitude. 
We provide additional details about Gated SAEs and their training in App.~\ref{app:sae}.

A key idea underlying SAE-based interpretability is that one can reinterpret the language model's (LM) internal computations in terms of the SAE features. By applying the decomposition in Eq.~\ref{eq:SAE-decomp} to hidden states $\vx$ in the LM and folding the SAEs into the forward pass,\footnote{As described in detail in \citet{marks2024sparsefeaturecircuitsdiscovering}.} we can express model representations as a combination of SAE feature activations $f_i$ and reconstruction errors $\vepsilon$.
This reframing enables gradient-based attribution patching techniques~\citep{syed2024attribution} to identify which SAE features have causal influence on particular aspects of model behavior.

\paragraph{Finding causally relevant features using attribution patching.}
Causal interpretability methods aim to locate components in neural networks~(such as attention heads or neurons) responsible for particular behaviors~\citep{mueller2024questrightmediatorhistory}.
This requires constructing a distribution $\mathcal{D}$ over pairs of inputs $(x_{\text{clean}}, x_{\text{patch}})$, where $x_{\text{clean}}$ is a prompt on which the behaviour occurs, and $x_{\text{patch}}$ is a reference prompt on which the behaviour does not occur.
Then, we can quantify the importance of components using a target metric $m: \mathbb{R}^{V} \rightarrow \mathbb{R}$, e.g., the difference in log-probabilities between a correct token continuation and minimally differing incorrect token continuation. 
In other words, we quantify the importance of a component on the model behaviour as the expected change in $\mathcal{L}$ when replacing the components's activation on the clean prompt with its value on the patch prompt.
This is known as the \textbf{indirect effect} (IE; \citealp{pearl-2001-indirect}).

An exact but compute-intensive method for causal analysis is activation patching~\citep{vig2020investigating}, where we observe the change in model behaviour when performing a counterfactual intervention to a component.
However, this requires a separate forward pass for each component.
Therefore, we use attribution patching~\citep[AtP;][]{kramar2024atpefficientscalablemethod} to estimate $\hat{\text{IE}}$, a linear approximation of the IE computed as a first-order Taylor expansion:
\begin{equation}
\begin{aligned}\label{eq:ig}
       \hat{\text{IE}}_\text{atp}&(m;\va;x_\text{clean}, x_\text{patch}) = \\ & \left.\nabla_{\va}m\right|_{\va = \va_\text{clean}}\left(\va_\text{patch} - \va_\text{clean}\right) 
\end{aligned}
\end{equation}
Here, $m$ is the target metric, usually defined as the logit difference between a correct and minimally differing incorrect token completion given context $x_\text{clean}$, $\va_\text{clean}$ is the activation of component $\va$ given $x_\text{clean}$, and $x_\text{patch}$ is a minimally differing context that changes the correct answer. Thus, the intuition is that we can approximate the causal contribution of $\va$ to $m$ by taking the slope of $m$ with respect to $\va$ (where this is approximated as the gradient of $m$ at $\va$\footnote{The gradient can be viewed as a local estimate of how much the metric we backpropagate from would be affected by changing the activation of the component.}) and multiplying this by the change in activations in $\va$.

AtP requires two forward passes and one backward pass to compute an estimate score for \emph{all} components in parallel. In practice, we employ a more expensive but more accurate approximation based on integrated gradients \citep{sundararajan2017axiomatic}.
For details, please see App.~\ref{app:at-ig}.

\paragraph{Probing classifiers.} 
\label{sec:probes}
Linear probes have frequently been used to locate representations of morphosyntactic features~\citep[e.g.,][]{hewitt-manning-2019-structural, giulianelli-etal-2018-hood,chi-etal-2020-finding}. Given labeled classification data, we may train logistic regressions to map from the activations of frozen intermediate layers of pre-trained models~\citep{belinkov2021probingclassifierspromisesshortcomings} to the task labels.
We similarly train classifiers to predict morphosyntactic concepts. We then quantify the extent of cross-lingual feature sharing for morphosyntactic concepts, and later evaluate the selectivity of targeted morphosyntactic concept interventions.

\begin{figure*}[!ht]
    \centering
    \includegraphics{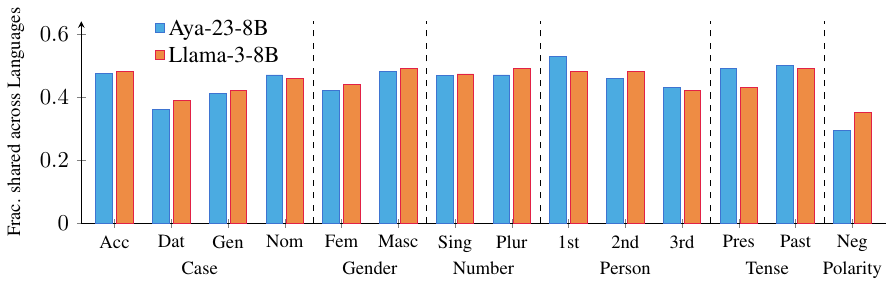}
    \vspace{-0.2cm}
    \caption{Proportion of features shared across languages (intersection over union) among the top 32 features for each morphosyntactic concept. 
    A significant fraction of the morphosyntactic concept representations are shared across languages in both Llama-3-8B and Aya-23-8B.}
    \label{fig:feature-overlap}
\end{figure*}

\section{Multilingual Features}
\label{cha:multilingual-features}
In this section, we investigate the extent to which morphosyntactic concepts are shared (or redundantly encoded) across languages.
We first measure the overlap between the most causally influential sparse features for a given concept across languages.
Then, we investigate whether ablating the multilingual features leads to a consistent decrease in classifier performance across languages.

\subsection{Experimental Setup}

\paragraph{Models.}
We consider Llama-3-8B~\citep{meta_llama_3} and Aya-23-8B~\citep{aryabumi2024aya}.
While Llama 3 was trained primarily on English data, Aya was explicitly developed to support 23 typologically diverse and resource-diverse languages.

\paragraph{Data.} 
We select 23 languages from Universal Dependencies 2.1~\citep[UD;][]{ud-2.1}, a multilingual treebank containing dependency-parsed sentences.
This corresponds to the 23 languages that Aya-23 was trained on (see App.~\ref{app:languages}).
Each word in each sentence in UD is annotated with its part of speech and morphosyntactic features, as defined in the UniMorph schema~\citep{kirov-etal-2018-unimorph}.

\paragraph{Training probing classifiers.}
\label{sec:probes}
We train a probing classifier for each combination of morphosyntactic concept (e.g., gender) and language (e.g., English).
We filter for morphosyntactic concepts that are annotated in at least six languages.
During training, the inputs to the classifiers are residual stream activations from the end of the model’s middle layer ($L=16$), as this layer is expected to capture abstract, high-level and human-interpretable features~\citep{elhage2022solu, lad2024remarkablerobustnessllmsstages}. 
Following~\citet{marks2024sparsefeaturecircuitsdiscovering}, we pool non-padding token activations by summing them; then, we fit a logistic regression to the pooled representations.

\paragraph{Training sparse autoencoders.}
We train Gated SAEs \citep{rajamanoharan2024improvingdictionarylearninggated} for both Llama-3-8B and Aya-23-8B. We collect activations over 250 million tokens of The Pile~\citep{gao2020pile} extracted from the residual stream at the same layer at which we train the probing classifiers.
To verify the quality of the SAEs, we measure the proportion of loss recovered when replacing the model activations with SAE reconstructions thereof. While the training dataset is primarily composed of English texts, we verify that it also reconstructs other languages effectively using the mC4 dataset (\citealp{xue-etal-2021-mt5}; see App.~\ref{app:sae}).

\subsection{Results}

\paragraph{Representations of morphosyntactic concepts are shared across languages.} 
First, we measure the extent of overlap across languages for the top features encoding a morphosyntactic category.
For this, we perform attribution patching (\S\ref{cha:methods}), where the target metric is the logit of the probing classifiers we trained in \S\ref{sec:probes}. This identifies the set of causally relevant features that are most influential on the probe's predictions.
We observe that a small fraction of features dominate the causal effect, with a long tail of features that have a small effect.
Therefore, for each language and concept, we select the top-32 most informative features per probe and compute the overlap across languages within a concept. 
We present the results in Figure~\ref{fig:feature-overlap}.

We observe significant overlap of up to 50\% in the features identified for the same grammatical concepts across typologically diverse languages.
For example, feature 22860 is among the most influential features across \emph{all} 15 languages that inflect for grammatical gender.
The overlap is greater for morphosyntactic concepts present in a larger number of languages, such as grammatical number.
We also find that the extent of overlap is largely consistent across Llama-3-8B and Aya-23-8B, with only a small number of exceptions (e.g. negative polarity).
Thus, despite the lack of a one-to-one mapping of grammatical concepts across languages, there exist highly multilingual representations of abstract morphosyntactic concepts.

\begin{figure*}[!t]
    \centering
    \resizebox{\linewidth}{!}{
    \input{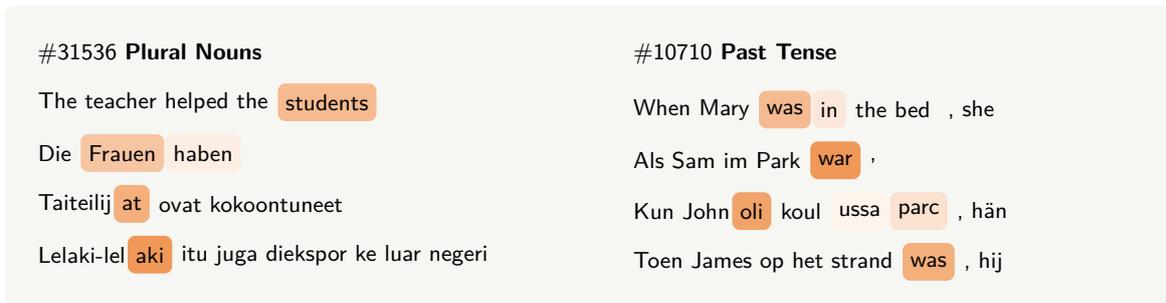}
    }
    \caption{Examples of the activation patterns of selected features in Llama-3-8B that correspond to cross-lingual representations of grammatical concepts. For example, we locate features that indicate the presence of plural nouns or features that indicate past tense across languages.}
    \label{fig:feature-activations}
\end{figure*}
\begin{figure}[!h]
    \centering
    \includegraphics{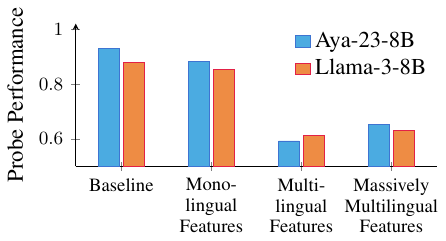}
    \vspace{-0.2cm}
    \caption{Performance of the probing classifiers before and after ablating features. Specifically, we test ablating (a) all monolingual features, (b) all multilingual features, and (c) the upper quartile of the \textit{most multilingual} features. We find that the classifiers crucialy rely on massively multilingual features to predict the presence of morphosyntactic features.}
    \label{fig:probe-feature-ablation}
    \vspace*{-0.4cm}
\end{figure}

\paragraph{Effect of causal interventions on multilingual features are consistent across languages.}
How important are multilingual features compared to monolingual features?
We evaluate the importance of multilingual features by measuring the impact of ablating the multilingual features on the performance of the probing classifiers.
We define multilingual features as those in the top feature set across at least two languages.
The results (Figure~\ref{fig:probe-feature-ablation}) suggest that the classifiers crucially rely on multilingual features to predict the presence of morphosyntactic concepts. This suggests that large language models---including those trained primarily on English---learn to rely on shared representations to detect particular concepts, rather than relying on language-specific representations.

We also observe that while many features are shared across at least two languages, there is a small set of features shared across a large number of languages (App.~\ref{cha:features_across_languages}).
Therefore, we investigate ablating only \textit{massively multilingual} features.
Specifically, we order features by the number of languages for which they are causally relevant, and the average strength of their causal effect. 
Then, we select the upper quartile of features in this ranking---i.e., the most multilingual features---and only ablate those.
For Llama 3, the performance after ablating only massively multilingual features drops to 64\% on average. 
Notably, when ablating \emph{all} multilingual features, the performance drops to 61\%---a difference of only 3\%, despite ablating four times as many features.
This suggests that the most massively multilingual features explain most of the probes’ behavior.

\paragraph{Feature overlap across concepts.} 
We now investigate the extent of feature overlap across grammatical concepts. 
This provides insights into the specificity of the functional roles of features. 
Specifically, we analyze the overlap in the sets of multilingual features associated with different concept-value pairs (e.g., masculine gender and singular number).\footnote{We use \emph{concept} to refer to the variable---e.g., gender or number---and \emph{concept-value} to refer to specific values that the variable can take---e.g., masculine gender or singular number.}
For each pair, we measure the fraction of multilingual features shared, and compute the mean overlap across all pairs.

We find that the mean overlap across concepts is 13.9\% ($\pm$ 10.7\%). 
This relatively low (but significant) overlap is expected, as some linguistic concepts are jointly morphologically realized. 
For instance, singular number and masculine gender share 40.6\% of their features; this is intuitive, given the frequent joint inflection of adjectives and pronouns for both number and gender. For example, in French and Hebrew, ``il'' and ``\textcjheb{'wh}'' are masculine singular pronouns, where changing either number \emph{or} gender alters the form.
Similarly, adjectives in these languages have distinct forms for combinations of grammatical gender and number.

\paragraph{Multilingual features are human-interpretable.}
One benefit of sparse features over neurons is their interpretability. This provides a qualitative way to verify that the features are truly relevant to the target concept, rather than related solely in some spurious correlational manner. Thus, we manually inspect a subset of the identified features, finding that many of them are intuitively meaningful and interpretable across different languages. 
We present the most multilingual features for two common concepts and a selection of their activation patterns across typologically diverse languages in Figure~\ref{fig:feature-activations}.

\section{Demonstrating Functional Selectivity: A Case Study in Machine Translation}
Another benefit of sparse features is the level of fine-grained control they provide over language model behaviors \citep{templeton2024scaling}. Indeed, feature-based interventions tend to be more effective then neuron-based methods at precisely modifying how language models generalize without destroying task performance \citep{marks2024sparsefeaturecircuitsdiscovering}. We leverage this property in a machine translation setting to (i) demonstrate the \textbf{functional selectivity} of the discovered morphosyntactic features (i.e., to show that intervening with them has few side effects on other concepts), and (ii) provide additional evidence that their causal role is highly \textbf{general} (i.e., that it is effective in this MT setting and not just on the probes). 
To evaluate (i) and (ii), we use feature steering~\citep{templeton2024scaling}, where we clamp features of interest to counterfactual activations. These activations can be either significantly higher or lower than the values observed in practice.
This approach is similar to prior work on understanding the functional role of individual neurons in image models~\citep{bau2018visualizing}.

\begin{figure*}[!t]
    \centering
    \includegraphics{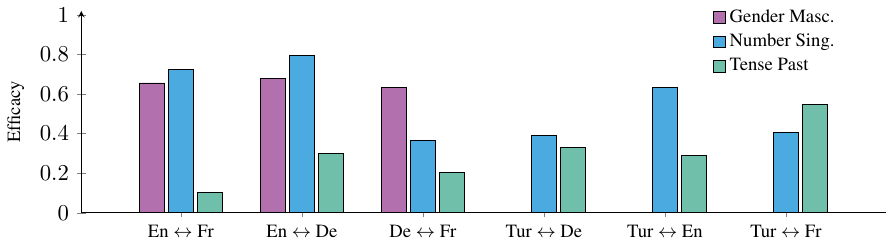}
    \vspace{-0.2cm}
    \caption{Efficacy in flipping the model behavior on our dataset when translating while intervening on a \textit{single} multilingual feature per concept. 
    For each concept, we translate a sentence containing some concept (e.g., present tense) from some source language to another language while intervening on a feature, and measure the number of times the model generates a translation containing the counterfactual concept value (e.g. past tense).
    In each setting, we intervene on a single feature and measure the success rate over 64 examples. Results aggregated across translation directions; see App.~\ref{app:mt-directional} for separate results per direction.}
    \label{fig:efficacy-label-flipping}
\end{figure*}

As translation is a difficult task that does not always have a single clear answer, we restrict to a smaller subsample of concepts and languages such that we can manually verify the baseline quality of translations, and then the efficacy of the interventions. We restrict our analysis to tense (Past vs.\ Present), gender (Masc.\ vs.\ Fem.), and number (Sing.\ vs.\ Plural), as these concepts are shared across most languages we consider. We analyze English, French, and German; we additionally include Turkish, but restrict our focus to number and tense here, as Turkish does not inflect for grammatical gender.
We provide details on the prompt format and translation performance in Appendix~\ref{app:mt-details}.

\subsection{Experimental Setup}

\paragraph{Data.}
We design datasets consisting of minimal pairs of inputs that differ only with respect to the presence of a grammatical concept (see Table~\ref{tab:counterfactual-examples} in App.~\ref{app:counterfactual-data}).
For example, we generate counterfactual pairs that elicit singular or plural verbs based on the grammatical number of the subject:
\begin{itemize}[noitemsep]
    \itemsep0em
    \item[a.] The \textbf{\textcolor{blue}{parents}} near the cars $\rightarrow$ \textbf{\textcolor{blue}{were}}
    \item[b.] The \textbf{\textcolor{red}{parent}} near the cars $\rightarrow$ \textbf{\textcolor{red}{was}}
\end{itemize}
This is an adaptation and translation of data from \citet{arora-etal-2024-causalgym}. Then, we check whether the probability of the grammatical sequence is higher for each pair. 
In our experiments, we only consider sentences where Llama-3-8B is capable of making grammatically correct predictions.
We find that it is almost always correct.

\paragraph{Feature intervention.}
We follow~\citet{templeton2024scaling} and intervene in the model by setting a single SAE feature to a multiple of its maximum activation value observed in the dataset.\footnote{For this purpose, we record the maximum activations of each feature across 20M tokens from mC4 \citep{xue-etal-2021-mt5}.}
Specifically, we decompose the residual stream activations $\vx$ into the sum of two components, the SAE reconstruction and the reconstruction error (see Eq.~\ref{eq:SAE-decomp}).
Then, we replace the SAE term with a modified SAE reconstruction in which we set the activation value of a specific feature to a counterfactual value, but leave the error term unchanged.
We then run the forward pass on the network using this modified residual stream.
We only apply the intervention to the last token position of each generation step to prevent degenerate outputs. 

To select the feature to intervene with, we manually inspect feature activation patterns. 
We select one feature per concept from the set of multilingual features identified in \S\ref{cha:multilingual-features} where the activation values are most interpretable.
Similar to~\citet{templeton2024scaling}, we explore different multiples and choose the one that works best empirically.

\subsection{Quantitative Results}
\paragraph{Efficacy.}
We define the efficacy of the intervention as the proportion of examples where the probe's label flips after the intervention. To measure this, we generate a translation with and without the intervention. Then, in new forward passes with no interventions, we give the model these translations, pool their representations, and measure whether the probe's (from \S\ref{sec:probes}) labels differ between the two translations. We find that intervening on a single feature is often sufficient to change the models prediction towards the indended concept (e.g., a masculine instead of feminine pronoun; see Figure~\ref{fig:efficacy-label-flipping}). 
However, for some concepts, intervening on a single feature is often insufficient.

\begin{figure}[!t]
    \centering
    \includegraphics{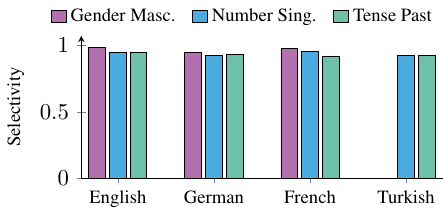}
    \vspace{-0.2cm}
    \caption{Fraction of times an intervention on a grammatical concept exclusively impacts that concept, indicating the degree of cross-concept interference. We find that interference is generally very rare. Note that Turkish does not inflect for grammatical gender.}
    \label{fig:selectivity}
\end{figure}

\begin{figure*}[!t]
    \centering
    \resizebox{\linewidth}{!}{
    \input{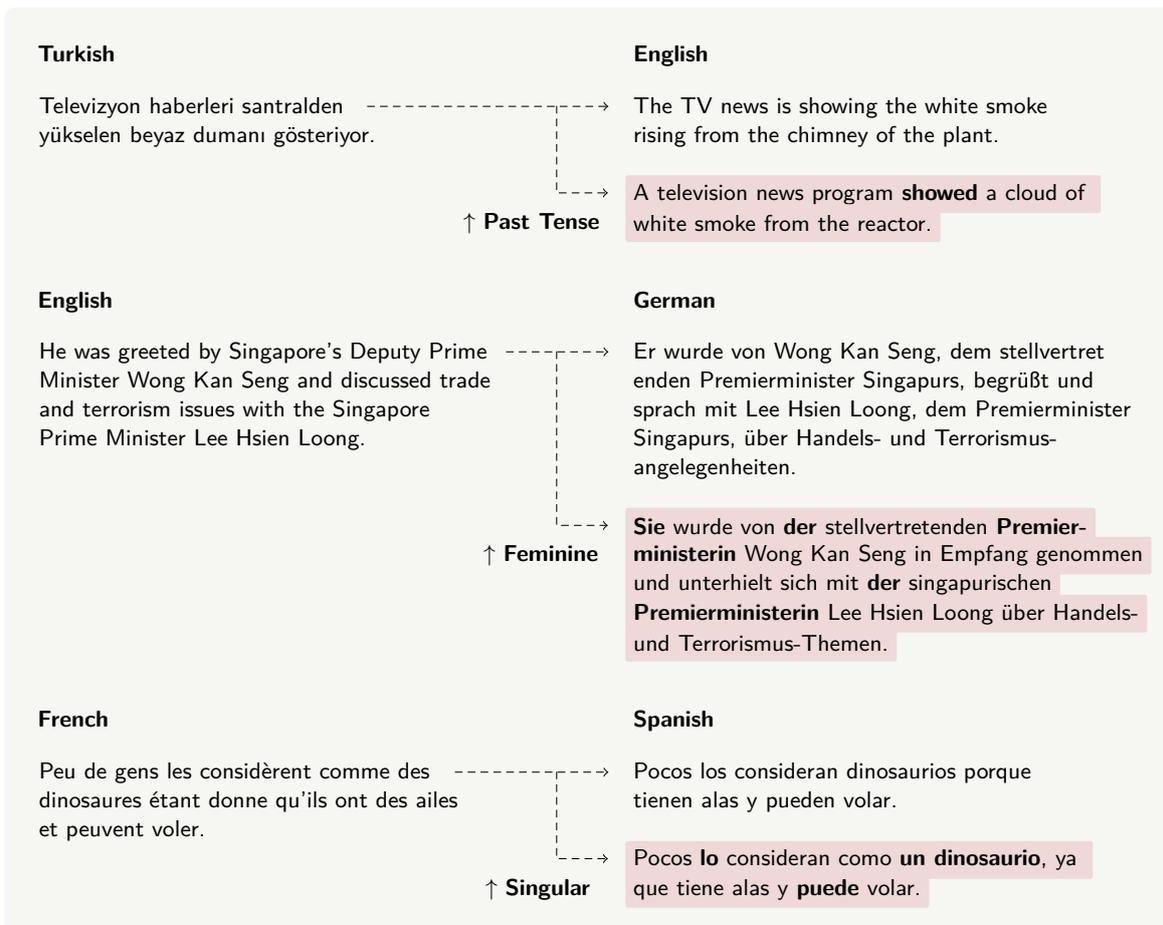}
    }
    \caption{Comparison of translations with (highlighted in \textcolor{redhighlight}{red}) and without intervening on multilingual features encoding a morphosyntactic concept. 
    The words indicating that the model flipped its behavior are highlighted in bold. 
    The input texts are sampled from the Flores-101 dataset.}
    \label{fig:translations-qualitative}
\end{figure*}

\paragraph{Selectivity.}
We also evaluate the selectivity of the interventions. 
Specifically, we want interventions to \emph{only} flip the labels on the concept that we intervene on, and not others.
We operationalise this using the probing classifiers trained in \S\ref{sec:probes}. As before, we generate two translations: one with and without feature interventions to a given concept. Then, we compute selectivity as the proportion of translation pairs for which none of the other concept probes changed their predictions. 
Our results (Figure~\ref{fig:selectivity}) suggest that the interventions are generally selective to the concept.

\subsection{Qualitative Results}
The quantitative results provide insights into the generalisation of the identified features to a different distribution, and allow us to quantify the success rate of the interventions.
However, the dataset is neither naturalistic nor diverse.
Therefore, we perform the same interventions on the Flores-101 dataset~\citep{flores101}, which provides aligned translations across 101 languages.
We present a selection of qualitative examples in Figure~\ref{fig:translations-qualitative}.

However, we acknowledge that the intervention was not successful in some cases, and sometimes led to degenerate generations. Concept- and language-specific tuning over the activation values was also required.
These limitations highlight common problems with feature steering methods, including that activations must often be scaled to significantly larger values than would naturally be observed~\citep{templeton2024scaling}. This, combined with intervening on multiple features simultaneously, can result in unexpected or degenerate outputs~\citep{lesswrong_sae_feature_Steering}.

\section{Related Work}
\paragraph{Multilingual language modeling.} Multilingual language models are trained on multiple languages simultaneously, typically with an explicitly designed balance between languages in the pretraining corpus. For example, XLM-R~\citep{conneau-etal-2020-unsupervised}, M2M~\citep{mohammadshahi-etal-2022-small}, mT5~\citep{xue-etal-2021-mt5}, and Aya~\citep{aryabumi2024aya} are trained with corpora balanced across many languages. In contrast, Llama 3 \citep{meta_llama_3} is trained on a much more English-centric distribution: over 90\% of the corpus is English text. Generally, the motivation behind balanced corpora during multilingual pretraining is to encourage cross-lingual transfer.
 We find that this balance may not be necessary for multilingual representations when the corpora are sufficiently large.

\paragraph{Interpreting multilingual language models.} A central question in multilingual language modeling is whether LMs develop universal concept representations, disentangled from specific languages. \citet{dhar-bisazza-2021-understanding} study different forms of cross-lingual transfer in recurrent neural networks and found that exposing our LMs to a related language does not always increase gram- matical knowledge in the target language. 
\citet{ruzzetti-etal-2023-exploring} study the similarity of monolingual encoder models, finding higher similarity between models trained for typologically similar languages.
\citet{wendler2024llamasworkenglishlatent} find that languages models trained on unbalanced, English-dominated corpora operate in a concept space that lies closer to English than to other languages. 
\citet{ferrando2024on} find that mechanisms of subject-verb agreement are consistent across languages. 
\citet{dumas2024how} observe that concepts (e.g., ``car'') are encoded disentangled from specific languages and can be transferred between them.
Similarly, \citet{stanczak-etal-2022-neurons} find that probes trained on different languages read the same concepts from the same neurons in smaller-scale masked language models such as mBERT \citep{devlin-etal-2019-bert} and XLM-R \citep{conneau-etal-2020-unsupervised}, while \citet{feng2024monitoring} observe that their propositional probes generalise to Spanish. 
\citet{varda-marelli-2023-data} extend this study by investigating the cross-lingual consistency of individual neurons responding to syntactic phenomena.
Our work extends these findings to large-scale models (including one not explicitly intended to be multilingual), and shows stronger causal evidence for the functional selectivity of these conceptual representations.

\section{Discussion and Conclusion}
Language models trained primarily on English data perform surprisingly well in other languages. Prior work has emphasized the importance of balanced pretraining corpora, so as to not overfit to a single language~\citep{conneau-etal-2020-unsupervised, mohammadshahi-etal-2022-small}; however, more recent studies suggest that the size of the pretraining corpus---not balance---may be more important for cross-lingual generalization \citep{jiang2023mistral7b,meta_llama_3,schafer2024language}. Why is this effective? Our results provide evidence that large-scale pretraining, even with imbalanced corpora, induces equally cross-lingualistically generalizable grammatical abstractions as pretraining with balanced corpora. This is evidence in favor of a hypothesis discussed in \citet{wendler2024llamasworkenglishlatent}: the internal \emph{lingua franca} of large language models may not be English words \emph{per se}, but rather concepts. 
That said, it is still reasonable to assume these concepts are likely biased toward how English handles them in models pre-trained on imbalanced corpora. 

How generalizable should grammatical concept representations be? While it is more parameter-efficient to share representations across languages, this may lead to biases if two languages assign different social or semantic connotations to---or simply distribute differently---the same grammatical concepts. Indeed, there is not a one-to-one relationship between concepts across languages: different languages may have differing numbers of values for the same concept (e.g., Finnish has many more grammatical cases than German), and may use the same categories in different ways (e.g., the same nouns often have different genders in different languages, even within language families). More language-specific features could lead to less bias, but would require more parameters to represent. It is not clear what the optimal point between these two ends of generalization are.

More practically, the extent to which concepts are shared across languages has implications for multilingual downstream NLP tasks. 
For example, it helps explain the phenomenon that preference tuning for toxicity mitigation in a single language generalizes to other languages~\citep{li2024preferencetuningtoxicitymitigation}.
Thus, we might reasonably expect various types of interventions and model editing approaches to generalize from a single language to other languages. %

\section*{Limitations}

\paragraph{Steering with SAEs.} 
We use SAE features to intervene on the models internal state. 
Similar to~\citet{templeton2024scaling}, we find that steering with SAEs typically requires clamping feature activations to values outside their observed range over the training dataset.
However, clamping feature activations to extreme values often causes the model to produce nonsensical generations, e.g., repeating the same token indefinitely. 
SAEs have also been shown to perform worse for fine-grained steering model behavior than alternatives~\citep{wu2025axbenchsteeringllmssimple}, such as difference-in-means~\citep{marks2024the}.
Thus, while our results demonstrate that the identified features can be causally responsible for model behavior, optimally steering model behavior is an open problem that future work should address.

\paragraph{Distinction between model understanding and generation.}
In our experiments, the linear probes measure whether it is possible to linearly separate grammatical concepts in the model's representations, whereas the machine translation experiments measure the effect of feature directions on model generation. Past work~\citep{meng2022locating, orgad2024llmsknowshowintrinsic, gottesman-geva-2024-estimating} has observed a distinction between a model encoding a concept versus being able to use that concept during generation (i.e., a ``knowing vs. saying'' distinction). Anecdotally, we observed that some features with a high causal effect on the linear probe do not necessarily have a similar effect on the model's generation. We hypothesize that this is due to an intrinsic distinction between concept understanding/representation and concept generation.

\paragraph{Human biases in feature interpretation.} One must be cautious in applying human concepts to LM representations. It is probable that LMs deploy distinct concept spaces from humans; thus, human explanations of neurons or features are likely to be biased in ways that can result in suboptimal predictions of when the neuron or feature will activate \citep[cf.][]{huang-etal-2023-rigorously}.

\paragraph{Non-linear features.} Sparse autoencoders are a method for unsupervised search for features. 
However, they will only recover a feature if it is \emph{linearly encoded}.
Thus, if a concept is encoded in a non-linear manner, then sparse features may struggle to recover the full range of that variable. 
Thus, sparse features are generally best-suited to encoding binary relations, whereas some features may be encoded in more complex arrangements such as circular shapes (cf.\ \citealt{engels2024language,csordas-etal-2024-recurrent}).

\section*{Acknowledgements}
Jannik Brinkmann is supported by the German Federal Ministry for Economic Affairs and a grant from the Long-Term Future Fund.
Chris Wendler is supported by a grant from Open Philanthropy.
Aaron Mueller is supported by a postdoctoral fellowship under the Zuckerman STEM Leadership Program.

\bibliography{acl_latex}

\appendix

\section{Implementation Details}
\subsection{Infrastructure}
The experiments were run on a single server with 8 NVIDIA RTX A6000 48 GB GPUs with CUDA Version 12.4 and an AMD EPYC 7413 24-Core Processor. 
The total runtime for training the probes and autoencoders was less than a week. 

\subsection{Libraries}
To extract activations and intervene on the model, we use \texttt{nnsight}~\citep{fiottokaufman2024nnsightndifdemocratizingaccess}. 
For training the linear probes, we use \texttt{scikit-learn}~\citep{scikit-learn}.
For training the SAEs, we use \texttt{dictionary-learning}~\citep{saprmarks_dictionary_learning}, itself based on \texttt{nnsight}.

\section{Languages}
\label{app:languages}

In our experiments, we consider the set of languages which Aya-23~\citep{aryabumi2024aya} was trained on. This includes: Arabic, Chinese, Czech, Dutch, English, French, German, Greek, Hebrew, Hindi, Indonesian, Italian, Japanese, Korean, Persian, Polish, Portuguese, Romanian, Russian, Spanish, Turkish, Ukrainian, and Vietnamese.

\section{Attribution Patching}
\label{app:at-ig}
Here, we elaborate on activation patching, attribution patching, and the modified attribution patching approach we use in our experiments.

\textit{Activation patching} is a method for quantifying the causal importance of a model component on a model behavior.
It involves replacing the activation of some component in the model during one model forward pass with the activations of the same component from a different forward pass.
This requires constructing a distribution $\mathcal{D}$ over pairs of inputs $(x_{\text{clean}}, x_{\text{patch}})$, where $x_{\text{clean}}$ is a prompt on which the behaviour occurs, and $x_{\text{patch}}$ is a reference prompt on which the behaviour does not occur.
For example, to study subject-verb agreement one could construct a datasets of pairs such as:
\begin{itemize}%
    \itemsep0em
    \item[a.] The \textbf{\textcolor{blue}{dog}} $\rightarrow$ \textbf{\textcolor{blue}{barks}}
    \item[b.] The \textbf{\textcolor{red}{dogs}} $\rightarrow$ \textbf{\textcolor{red}{bark}}
\end{itemize}
Then, we can define the contribution of a node $n \in N$, where $N$ is the number of components in the model, to the model's behaviour as the counterfactual expected impact of replacing that node on the clean prompt with its value on the patch prompt~\citep{meng2022locating}. In practice, performing activation patching for all components requires $\mathcal{O}(N)$ forward passes, which does not scale efficiently~\citep{mueller2024questrightmediatorhistory}. 
Therefore, we use a faster linear approximation of the contribution of a node; this is \textit{attribution patching}, which computes a first-order Taylor expansion of this metric as described in \S\ref{cha:methods}.

In our experiments, we use the attribution patching approach proposed by~\citet{marks2024sparsefeaturecircuitsdiscovering} which employs a more expensive but more accurate approximation based on integrated gradients: 
\begin{equation}
\begin{aligned}\label{eq:ig}
       &\hat{\text{IE}}_\text{ig}(m;\va_\text{clean},\va_\text{patch}) = \\
       &\left(\sum_{\alpha} \left.\nabla_{\va}m\right|_{\alpha \va_\text{clean} + (1 - \alpha)\va_\text{patch}}\right)(\va_\text{patch} - \va_\text{clean}) 
\end{aligned}
\end{equation}
where the sum in (\ref{eq:ig}) ranges over $K=10$ equally-equally spaced $\alpha\in \{0,\tfrac{1}{K},\dots,\tfrac{K-1}{K}\}$. 
While this cannot be done in parallel for two nodes when one is downstream of another, it can be done in parallel for arbitrarily many nodes which do not depend on each other. 
Thus the additional cost of computing $\hat{\text{IE}}_\text{ig}$ over $\hat{\text{IE}}_\text{atp}$ scales linearly in $K$.

\section{Sparse Autoencoders}
\label{app:sae}

\subsection{Gated Sparse Autoencoders}
The Gated SAE architecture, as introduced in \citet{bricken2023monosemanticity} is defined by encoder weights \( W_e \in \mathbb{R}^{m \times n} \), decoder weights \( W_d \in \mathbb{R}^{n \times m} \) with columns constrained to have a $L_2$-norm of 1, and biases \( b_e \in \mathbb{R}^{m} \), \( b_d \in \mathbb{R}^{n} \). Given an input $\mathbf{x}\in \mathbb{R}^n$, the SAE computes
\begin{align}
    \mathbf{f}(\mathbf{x}) &= \text{ReLU}(W_e (\mathbf{x} - \mathbf{b}_d) + \mathbf{b}_e) \\
    \mathbf{\hat{x}} &= W_d \,\mathbf{f}(\mathbf{x}) + \mathbf{b}_d
\end{align}
where \( \mathbf{f}(\mathbf{x}) \) is the vector of feature activations, and  \( \mathbf{\hat{x}} \) is the reconstruction. 
This model is trained using the following loss term: 
\begin{equation}
\mathcal{L}_\textrm{standard} = \mathbb{E}_{\mathbf{x}\sim\mathcal{D}_\textrm{train}}\Big[ \| \mathbf{x} - \mathbf{\hat{x}} \|_2 + \lambda \| \mathbf{f}(\mathbf{x}) \|_1\Big].
\label{eq:Lstandard}
\end{equation}
for some hyperparameter $\lambda>0$ controlling sparsity.

The $L_1$ penalty introduces biases that can harm the accuracy of the reconstruction, as the loss can be decreased by trading-off reconstruction for lower L1~\cite{wright2024suppression}. 
To address this, \citet{rajamanoharan2024improvingdictionarylearninggated} introduced a architectural modification that separates (i) the selection of dictionary elements to use in a reconstruction, and (ii) estimating the coefficients of these elements. This results in the following gated architecture: 

\vspace{-1em}
\[
\pi_\text{gate}(\mathbf{x}) := W_\text{gate}(\mathbf{x} - \mathbf{b}_{d}) + \mathbf{b}_\text{gate}
\]
\[
{\mathbf{\tilde{f}}}(\mathbf{x}) := \mathbb{I}\left[{\pi_\text{gate}}(\mathbf{x}) > 0\right] \odot \]
\[\text{ReLU}(W_\text{mag}(\mathbf{x} - \mathbf{b}_{d}) + \mathbf{b}_\text{mag})
\]
\[
{\hat{x}}({\mathbf{\tilde{f}}}(\mathbf{x})) = W_d {\mathbf{\tilde{f}}}(\mathbf{x}) + \mathbf{b}_d
\]
where \(\mathbb{I}[ \cdot > 0]\) is the Heaviside step function and \(\odot\) denotes elementwise multiplication. Then, the loss function uses \({\hat{x}_\text{frozen}}\), a frozen copy of the decoder:

\vspace{-1em}
\begin{align}
    \mathcal{L}_\text{gated} &:= \mathbb{E}_{\mathbf{x}\sim \mathcal{D}_\textrm{train}} \Big[  \|\mathbf{x} - \hat{x}(\mathbf{\tilde{f}}(\mathbf{x}))\|_2^2 \notag\\
    & + \lambda \|\text{ReLU}(\pi_\text{gate}(\mathbf{x}))\|_1 \label{eq:Lgated} \\
    & + \|\mathbf{x} - \hat{x}_\text{frozen}(\text{ReLU}(\pi_\text{gate}(\mathbf{x})))\|_2^2 \Big] \notag
\end{align}
\vspace{-2em}

\subsection{Training Parameters}

\begin{table}[h!]
    \centering
    \begin{tabular}{l c}
        \toprule
        \textbf{Parameter} & \textbf{Value} \\
        \midrule
         Optimiser & \texttt{Adam} \\
         Learning Rate & $0.001$ \\
         L1 Coefficient & $0.505$ \\
         Expansion Factor & $8$ \\
         Number of Token & $250$ Million \\
         Batch Size & $512$ \\
         Warmup Steps & $1,000$ \\
         \bottomrule
    \end{tabular}
    \caption{Training parameters of our sparse autoencoders for both Llama-3-8B and Aya-23-8B.}
    \label{tab:sae-training-parameters}
\end{table}

\subsection{Performance Across Languages}
We evaluate the performance of the SAE trained on Llama-3-8B using the multilingual C4 (mC4) dataset \citep{xue-etal-2021-mt5}, itself a multilingual extension of the C4 dataset \citep{2019t5}. 
Table \ref{tab:sae_loss_recovered} presents the \textit{loss recovered} when replacing the residual stream activations with the SAE reconstructions across a range of languages. 
For this purpose, we compute the loss recovered as 
\begin{equation}
    \mathcal{L}_\text{recovered} = \frac{\mathcal{L}_\text{reconstructed} - \mathcal{L}_\text{zero}}{\mathcal{L}_\text{original} - \mathcal{L}_\text{zero}}
\end{equation}
where $\mathcal{L}_\text{original}$ is the cross entropy (CE) loss of the language model without intervention, $\mathcal{L}_\text{zero}$ is the CE loss of the model when zero ablating the activations, and $\mathcal{L}_\text{reconstructed}$ is the loss when replacing the activations with the SAE reconstructions.
This is a standard extrinsic metric to measure autoencoder quality~\citep{mueller2024questrightmediatorhistory}. 
However, it does not measure the intrinsic interpretability of the SAE features, which is not possible to measure in the absence of ground truth features~\citep{karvonen2024measuring, makelov2024towards}.

\begin{table}[ht!]
    \centering
    \begin{tabular}{c c}
         \toprule
         Language & Loss Recovered \\
         \midrule
         Chinese & 0.8122\\
         Dutch & 0.9004\\
         English & 0.9257 \\
         Finnish & 0.8582 \\
         French & 0.9285 \\
         German & 0.9026 \\
         Indonesian & 0.9205 \\
         Japanese & 0.8759 \\
         Turkish & 0.8522 \\
         \bottomrule
    \end{tabular}
    \caption{Evaluation of the sparse autoencoder trained on Llama-3-8B across different subsets on the AllenAI C4 dataset. The table shows the fraction of loss recovered when replacing residual stream activations with sparse autoencoders reconstructions across nine languages.}
    \label{tab:sae_loss_recovered}
\end{table}

\section{Features Across Languages}
\label{cha:features_across_languages}

\begin{figure}[!h]
    \centering
    \includegraphics{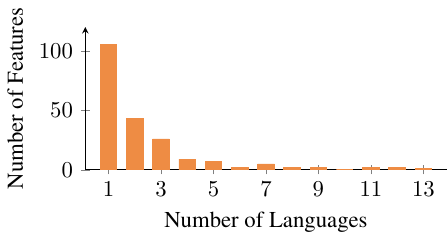}
    \vspace{-0.3cm}
    \caption{Distribution of the number of languages across which a given feature associated with masculine gender is shared (Llama 3).}
    \vspace{-0.4cm}
    \label{fig:languages_per_feature_gender_masc_llama}
\end{figure}

\begin{figure}[!h]
    \centering
    \includegraphics{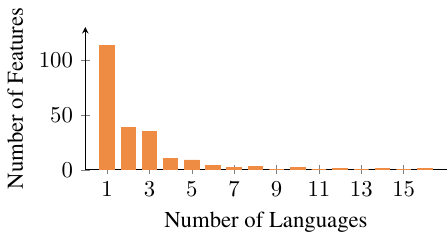}
    \vspace{-0.3cm}
    \caption{Distribution of the number of languages across which a given feature associated with past tense is shared (Llama 3).}
    \vspace{-0.4cm}
    \label{fig:languages_per_feature_tense_past_llama}
\end{figure}

\begin{figure}[!h]
    \centering
    \includegraphics{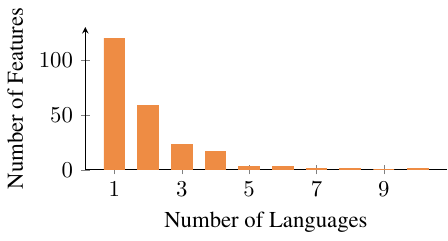}
    \vspace{-0.3cm}
    \caption{Distribution of the number of languages across which a given feature associated with accusative case is shared (Llama 3).}
    \label{fig:languages_per_feature_case_acc_llama}
\end{figure}

We measured the number of languages for which the top-32 features for each concept were strongly influential. 
Across grammatical concepts, the number of languages that a feature is influential for roughly follows a power law distribution. 
Below, we present the distributions for a selection of common concepts (see Figures~\ref{fig:languages_per_feature_gender_masc_llama}, \ref{fig:languages_per_feature_tense_past_llama}, \ref{fig:languages_per_feature_case_acc_llama}).
We find that for most concepts, there are multiple features that are shared across nearly all languages.
Note that the maximum number of languages varies across concepts, as not all languages inflect for a given concept nor are annotated for it in Universal Dependencies.


\begin{figure}[!h]
    \centering
    \includegraphics{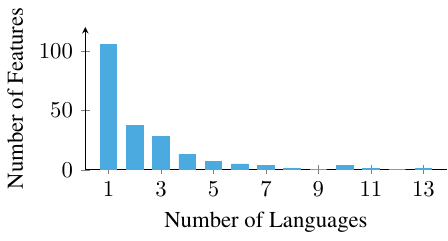}
    \caption{Distribution of the number of languages across which a given feature associated with masculine gender is shared (Aya-23).}
    \vspace{-0.2cm}
    \label{fig:languages_per_feature_gender_masc_llama_aya}
\end{figure}
\begin{figure}[!h]
    \centering
    \includegraphics{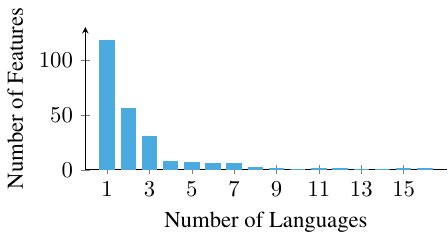}
    \caption{Distribution of the number of languages across which a given feature associated with past tense is shared (Aya-23).}
    \vspace{-0.2cm}
    \label{fig:languages_per_feature_tense_past_llama_aya}
\end{figure}
\begin{figure}[!h]
    \centering
    \includegraphics{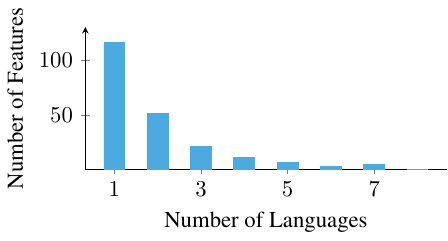}
    \caption{Distribution of the number of languages across which a given feature associated with accusative case is shared (Aya-23).}
    \label{fig:languages_per_feature_case_acc_llama_aya}
\end{figure}

\newpage
\section{Machine Translation Details}
\label{app:mt-details}

\subsection{Prompt Format}\label{app:mt-prompt}

We present an example of our 2-shot translation prompt in Figure~\ref{fig:translation_examples}.

\subsection{Translation Performance}\label{app:mt-performance}
\begin{figure}[h!]
    \centering
    \includegraphics[width=0.95\linewidth]{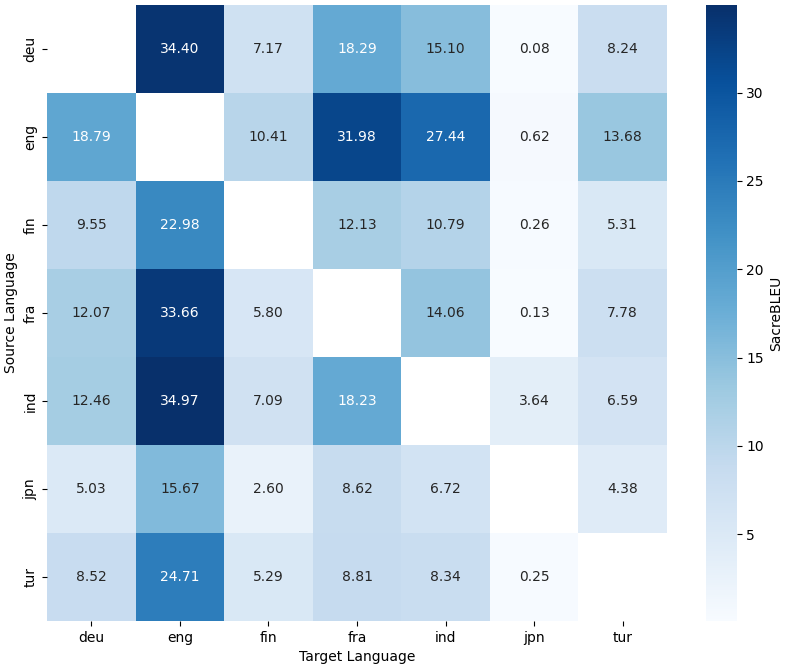}
    \caption{SacreBLEU scores of Llama-3-8B.}
    \label{fig:translation-bleu}
\end{figure}

\begin{figure}[h!]
    \centering
    \includegraphics[width=0.95\linewidth]{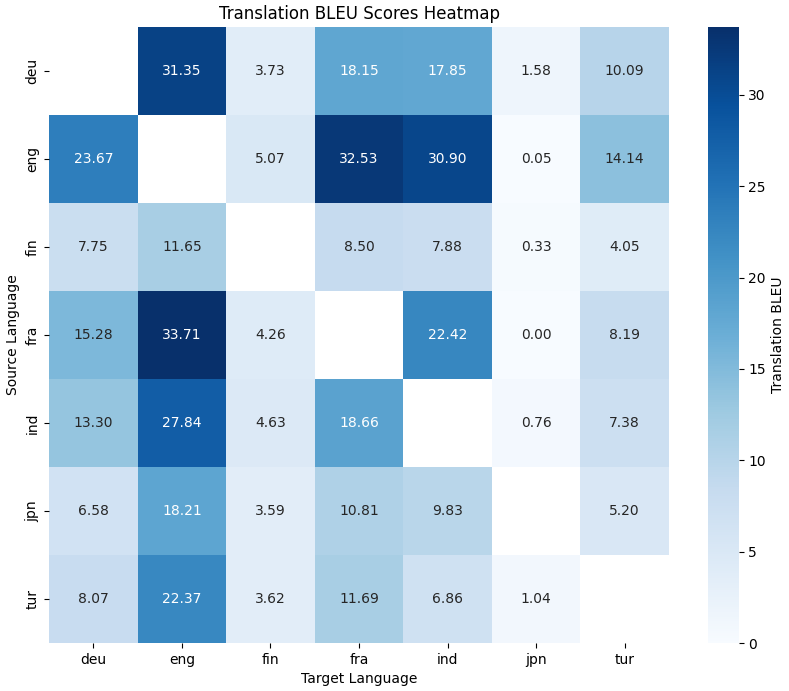}
    \caption{SacreBLEU scores of Aya-23-8B.}
    \label{fig:translation-bleu-aya}
\end{figure}

We evaluate the ability of Llama-3-8B and Aya-23-8B to translate languages in-context using the Flores-101 dataset~\citep{flores101}. 
Specifically, we measure the BLEU score~\citep{papineni-etal-2002-bleu} using the SacreBLEU implementation~\citep{post-2018-call}. We prompt models with 2 ground-truth exemplar translations, and then give it the test sentence in the source language (see Figure~\ref{fig:translation_examples}).
Results are presented in Figure~\ref{fig:translation-bleu} and \ref{fig:translation-bleu-aya}.

\begin{figure}[ht]
    \noindent\texttt{The governor's office said nineteen of the injured were police officers.} \textbackslash\textbackslash \;
    \texttt{Das Gouverneursamt erklärte, dass von den Verletzten neun Polizisten waren.}
    \vspace{0.2cm}
    
    \noindent\texttt{He produced over 1,000 stamps for Sweden and 28 other countries.} \textbackslash\textbackslash \;
    \texttt{Er produzierte über 1.000 Briefmarken für Schweden und 28 weitere Länder.} 
    \vspace{0.2cm}
    
    \noindent\texttt{The truck driver, who is aged 64, was not injured in the crash.} \textbackslash\textbackslash \;
    \caption{Example 2-shot translation prompt between English and German. Note that the double backslashes are \emph{literals}, not newlines. We found this prompt format to work best empirically in 2-shot settings in initial experiments.}
    \label{fig:translation_examples}
\end{figure}

\begin{figure*}[!t]
    \centering
    \includegraphics{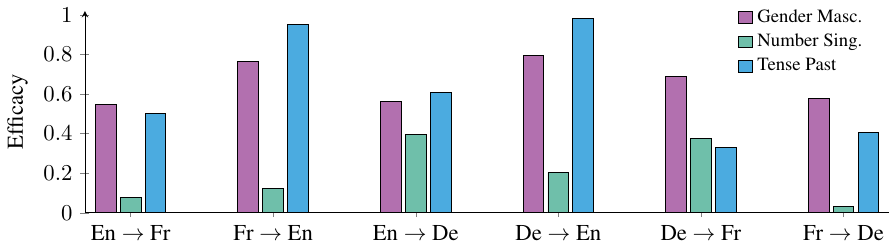}
    \vspace{-0.2cm}
\end{figure*}

\begin{figure*}[!t]
    \centering
    \includegraphics{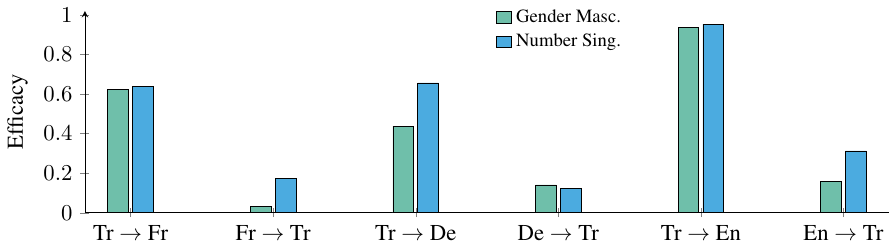}
    \vspace{-0.2cm}
    \caption{Efficacy in flipping the model behavior on our counterfactual dataset when translating between languages and intervening on a \textit{single} multilingual feature per concept. 
    For each concept, we translate a sentence from some source language (e.g. German) where some concept (e.g. present tense) to another language (e.g. English) and measure the number of times the model starts to predict some alternative concept value (e.g. past tense).
    In each of these settings, we intervene on a single feature and measure the success rate over 64 examples.}
    \label{fig:efficacy-label-flipping-appendix}
\end{figure*}

\subsection{Directional Translation Results}\label{app:mt-directional}
In Fig.~\ref{fig:efficacy-label-flipping}, we present averages for language pairs translating in both directions (e.g., English to French and French to English). Here, we provide additional details about the results for each translation direction separately (see Fig.~\ref{fig:efficacy-label-flipping-appendix}). 

\newpage
\begin{table*}[t!]
    \centering
    \begin{tabular}{llll}
        \toprule
        \textbf{Morphosyntactic Concept} & \textbf{Language} & \multicolumn{1}{l}{\textbf{Example Input}} & \multicolumn{1}{l}{\textbf{Example Output}} \\
        \midrule
        
        Gender (Masc. vs. Fem.) & English & When \textbf{\textcolor{blue}{Sam}} was at the park & \textbf{\textcolor{blue}{he}} \\
        & & When \textbf{\textcolor{red}{Sarah}} was at the park & \textbf{\textcolor{red}{she}} \\
        & French & \textbf{\textcolor{blue}{Louis}} a accepté, parce & \textbf{\textcolor{blue}{qu’il}} \\
        & & \textbf{\textcolor{red}{Charlotte}} a accepté, parce & \textbf{\textcolor{red}{qu’elle}} \\
        & German & \textbf{\textcolor{blue}{Lukas}} lachte, weil & \textbf{\textcolor{blue}{er}} \\
        & & \textbf{\textcolor{red}{Hannah}} lachte, weil & \textbf{\textcolor{red}{sie}} \\
        
        Number (Sing. vs. Plur.) & English & The \textbf{\textcolor{blue}{scientist}} & \textbf{\textcolor{blue}{discovers}} \\
        & & The \textbf{\textcolor{red}{scientists}} & \textbf{\textcolor{red}{discover}} \\
        & French & \textbf{\textcolor{blue}{Le docteur}} & \textbf{\textcolor{blue}{est}} \\
        & & \textbf{\textcolor{red}{Les docteurs}} & \textbf{\textcolor{red}{sont}} \\
        & German & \textbf{\textcolor{blue}{Der Schauspieler}} & \textbf{\textcolor{blue}{ist}} \\
        & & \textbf{\textcolor{red}{Die Schauspieler}} & \textbf{\textcolor{red}{sind}} \\
        & Turkish & \textbf{\textcolor{blue}{Hemşire}} sonraki adımlardan emin & \textbf{\textcolor{blue}{değil}} \\
        & & \textbf{\textcolor{red}{Hemşireler}} sonraki adımlardan emin & \textbf{\textcolor{red}{değiller}} \\

        Tense (Pres. vs. Past) & English & When Mary \textbf{\textcolor{blue}{is}} in the bed, she & \textbf{\textcolor{blue}{sleeps}} \\
        & & When Mary \textbf{\textcolor{red}{was}} in the bed, she & \textbf{\textcolor{red}{slept}} \\
        & French & Louis \textbf{\textcolor{blue}{marche} }parce qu'il & \textbf{\textcolor{blue}{est}}  \\
        & & Louis \textbf{\textcolor{red}{marchait}} parce qu'il & \textbf{\textcolor{red}{était}} \\
        & German & Lukas \textbf{\textcolor{blue}{geht}}, weil er & \textbf{\textcolor{blue}{ist}} \\
        & & Lukas \textbf{\textcolor{red}{ging}}, weil er & \textbf{\textcolor{red}{war}} \\
        & Turkish & Hasan \textbf{\textcolor{blue}{arıyor}} çünkü o & \textbf{\textcolor{blue}{hastadır}} \\
        & & Hasan \textbf{\textcolor{red}{aradı}} çünkü o & \textbf{\textcolor{red}{hastaydı}} \\
        \bottomrule
    \end{tabular}
    \caption{Examples from our counterfactual dataset.}
    \label{tab:counterfactual-examples}
\end{table*}

\section{Counterfactual Dataset}\label{app:counterfactual-data}
Our counterfactual dataset considers three concepts (gender, number, tense) and three languages (English, French, German). 
The dataset format is inspired by CausalGym~\citep{arora-etal-2024-causalgym} and the templates are in part taken from that dataset.
We present example counterfactuals in Table~\ref{tab:counterfactual-examples}.

\end{document}